\documentclass[letterpaper, 10 pt, conference]{ieeeconf}  

\IEEEoverridecommandlockouts                              

\usepackage{cite}
\usepackage{graphics} 
\usepackage{amsmath} 
\usepackage{amssymb}  
\usepackage{hyperref}
\usepackage{graphicx}
\usepackage{amssymb}
\usepackage{mathrsfs}
\usepackage{multirow}
\usepackage[table,xcdraw]{xcolor}
\usepackage{color}
\usepackage{subcaption}

\newcommand{\RomanNumeralCaps}[1]
    {\MakeUppercase{\romannumeral #1}}

\title{\LARGE \bf
Learning a Domain-Agnostic Visual Representation for \\Autonomous Driving via Contrastive Loss
}

\author{Dongseok Shim and H. Jin Kim$^{*}$

\thanks{This work was supported by Institute for Information \& Communications Technology Planning
\& Evaluation (IITP) grant funded by the Korea government (MSIT) under Grant
2019-0-01367, Infant-Mimic Neurocognitive Developmental Machine Learning
from Interaction Experience with Real World (BabyMind).}%
\thanks{Authors are with the Department
of Mechanical and Aerospace Engineering, Seoul National University,
Gwanak-gu, Seoul, 08826, Korea. E-mail: 
        {\tt\small \{tlaehdtjr01, hjinkim\}@snu.ac.kr} ($^{*}$Corresponding author)}%
}

\begin{document}

\maketitle
\thispagestyle{empty}
\pagestyle{empty}

\begin{abstract}

Deep neural networks have been widely studied in autonomous driving applications such as semantic segmentation or depth estimation. 
However, training a neural network in a supervised manner requires a large amount of annotated labels which are expensive and time-consuming to collect. Recent studies leverage synthetic data collected from a virtual environment which are much easier to acquire and more accurate compared to data from the real world, but they usually suffer from poor generalization due to the inherent domain shift problem. 
In this paper, we propose a Domain-Agnostic Contrastive Learning (DACL) which is a two-stage unsupervised domain adaptation framework with cyclic adversarial training and contrastive loss. DACL leads the neural network to learn domain-agnostic representation to overcome performance degradation when there exists a difference between training and test data distribution.
Our proposed approach achieves better performance in the monocular depth estimation task compared to previous state-of-the-art methods and also shows effectiveness in the semantic segmentation task. The official implementation of the paper in PyTorch is available at \texttt{\href{https://github.com/dsshim0125/dacl}{https://github.com/dsshim0125/dacl}}.

\end{abstract}

\section{INTRODUCTION}

Deep convolutional neural network (ConvNet) has led to the exceptional development of computer vision across a variety of tasks such as object detection\cite{he2017mask, redmon2016you}, semantic segmentation\cite{ronneberger2015u, chen2017rethinking}, and depth estimation\cite{godard2017unsupervised, fu2018deep}.
Most training algorithms with a supervised setup require a large amount of data as the performance of the ConvNet usually depends on the size of the training data and the quality of accurate labels. 

Unfortunately, it takes an intense and expensive effort to collect labels in the real world, especially in autonomous driving setting, since the data collected from the sensors are usually imperfect for training ConvNet and require an additional process. 
RGB-D camera like Kinect exhibits noisy artifacts\cite{song2015sun} that need an error correction for an accurate ground-truth depth map and the output of LiDAR is sparse that requires an extra interpolation to generate densely annotated labels.

To avoid these undesired issues, recent studies utilize computer graphics-based virtual environment to easily generate synthetic data paired with accurate task-specific labels \cite{richter2016playing, song2016ssc, Gaidon:Virtual:CVPR2016}. The problem is that ConvNet trained with synthetic data cannot perform well when evaluated on the real world data due to a domain gap between two different environments.

\begin{figure}[t]
    \centering
    \includegraphics[width=0.5 \textwidth]{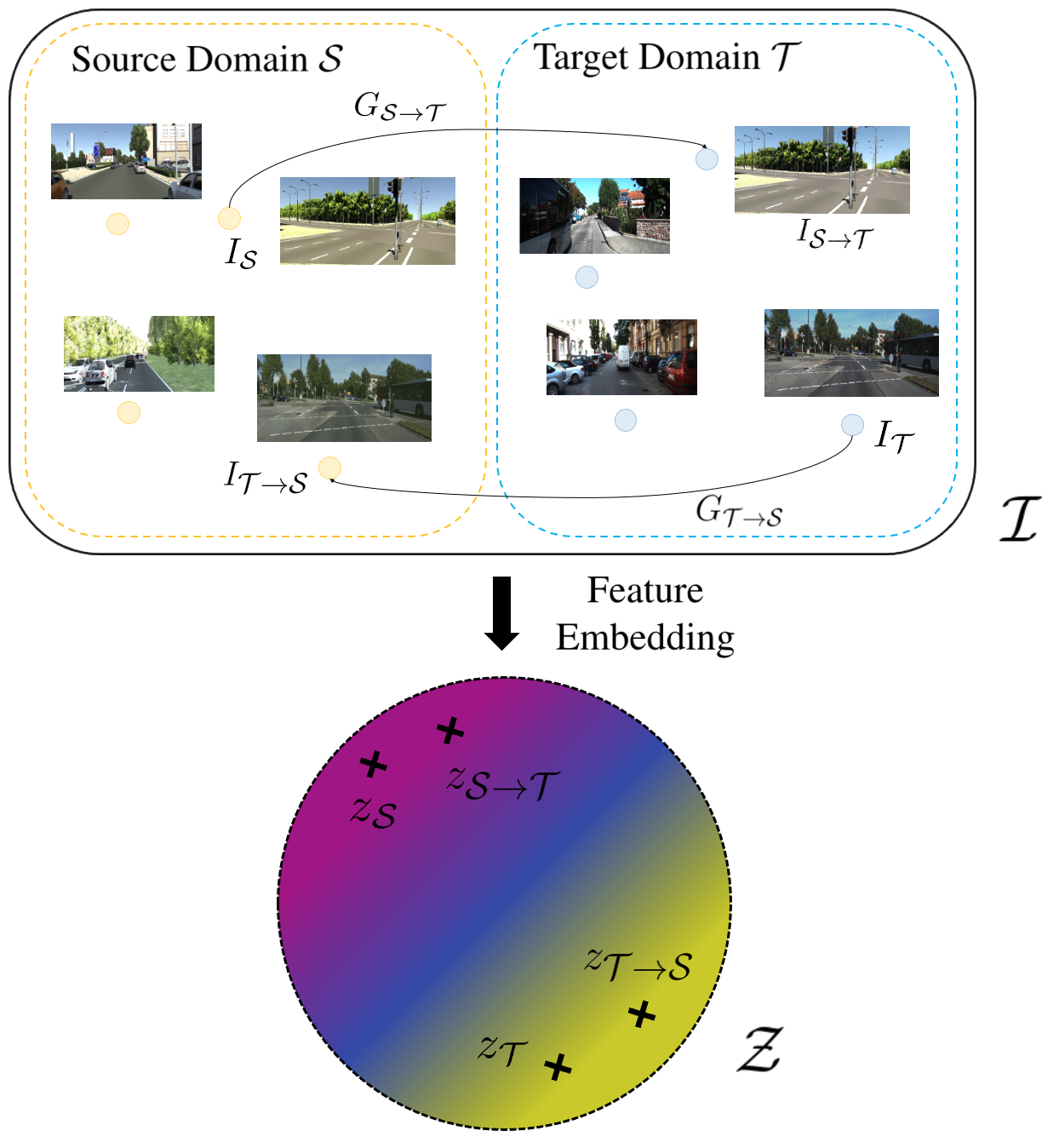}
    \caption{Two bidirectional style transfer networks, $G_{\mathcal{S}\rightarrow\mathcal{T}}$ and $G_{\mathcal{T}\rightarrow\mathcal{S}}$, map the source image $I_{\mathcal{S}}$ to the target domain $\mathcal{T}$ and the target image $I_{\mathcal{T}}$ to the source domain $\mathcal{S}$ respectively. 
    Then, a feature extractor followed by a projection head encodes the images from the image space $\mathcal{I}$ to the low-dimensional embedding space $\mathcal{Z}$. In the embedding space, the representation from the source image $z_{\mathcal{S}}$ and the representation from its fake target image $z_{\mathcal{S}\rightarrow\mathcal{T}}$ becomes similar and vise versa by our proposed domain-agnostic contrastive learning.}
    \label{fig:thumbnail}
\end{figure}

To mitigate such performance degradation, we propose a \textbf{Domain-Agnostic Contrastive Learning (DACL)} which employs momentum contrastive learning \cite{he2020momentum} with a GAN\cite{goodfellow2014generative}-based style transfer algorithm to learn the visual representation that is consistent and robust to domain shift. Our proposed DACL is composed of two stages which are style transfer stage and representation learning stage.

As DACL is an unsupervised domain adaptation framework, any supervisory cues, such as paired realistic images of the source data or task-specific labels, from the target domain do not exist. 
Instead, we only leverage unpaired images both from the source and the target domains to learn the style transfer function in a bidirectional way. 
The output of the function is used as the query data for the contrastive learning to train a feature extractor which learns the domain-agnostic representation of the image. 
Later, this trained feature extractor is used as the encoder of the task-specific network and the entire task network is finetuned with the densely annotated labels from the source domain in a supervised fashion.

Since the ConvNet is trained to learn the visual representation which contains domain-agnostic properties such as structural and semantic information, our method achieves better performance in monocular depth estimation and semantic segmentation compared to previous domain adaptation approaches. 

In short, our main contribution can be summarized as follows:
\\
\begin{itemize}
    \item We propose a two-stage unsupervised domain adaptation framework DACL which allows the ConvNet to extract domain-agnostic visual representation and be robust to domain shift.
    \item We show that contrastive representation learning with the adversarial style transfer function boost the performance of ConvNet for domain adaptation both in qualitative and quantitative metrics.
    \item We demonstrate that DACL outperforms the previous state-of-the-art depth estimation approaches with domain adaptation and also show its effectiveness on semantic segmentation task.\\
\end{itemize}

We evaluate our proposed method with two different tasks, monocular depth estimation and semantic segmentation. For both tasks, we leverage Virtual KITTI (vKITTI)\cite{Gaidon:Virtual:CVPR2016} as the synthetic source domain dataset and KITTI\cite{Menze2015CVPR} as the realistic target domain dataset.

\section{Related Work}

In this section, we summarize the past research on style transfer algorithms, visual representation learning with contrastive loss and recap domain adaptation strategies for several tasks. 

\subsection{Style Transfer}

Recent image-to-image translation or style transfer algorithms are mostly based on adversarial min-max optimization proposed in Generative Adversarial Network (GAN)\cite{goodfellow2014generative}.
Pix2Pix\cite{isola2017image} harnesses paired source and target datasets with conditional discriminator to translate the image distribution. 
However, in practice, it is difficult to collect paired images from different domains, and several works have explored unpaired image-to-image translation. 
CoGAN\cite{liu2016coupled} suggests training two identical networks with a weight-sharing strategy to learn the joint representation between two different domains, and CycleGAN\cite{zhu2017unpaired} introduces a cyclic consistency framework that consists of not only a source-to-target network but also a target-to-source network.
A cyclic reconstruction loss suggested in \cite{zhu2017unpaired} helps the style transfer network to preserve the structural information of the source image. 
Later, DiscoGAN\cite{kim2017learning} suggests bidirectional cyclic consistency to be more robust to structural collapse after style transfer.

\subsection{Contrastive Learning}

Representation learning with contrastive loss has been widely developed since CPCv2\cite{henaff2020data} demonstrated a remarkable result in transfer learning, as the unsupervised pre-training with contrastive loss firstly surpassed supervised pre-training for object detection.
Contrastive learning promotes the neural network to learn the similar representation between positive pairs by maximizing the lower bound of the mutual information (MI). 
As the lower bound of MI also depends on the batch size of the data on a log scale, contrastive learning requires a large number of batch size during training. 
To alleviate this restriction, MoCo\cite{he2020momentum} suggests a dynamic dictionary with momentum-based training and SimCLR\cite{chen2020simple} shows that strong data-augmentation and the usage of projection head improve the performance most.
Our proposed DACL adopts momentum-based contrastive learning\cite{he2020momentum} and the projection head suggested in \cite{chen2020simple}  with the style transfer network to learn the domain-agnostic representation.

\subsection{Task-specific Domain Adaptation}
An objective of domain adaptation is to handle the failure in generalization when the model is trained on biased data distribution. 
Previous studies mostly focused on capturing the domain invariant features by minimizing Maximum Mean Discrepancy (MMD) \cite{long2015learning, gretton2012kernel}, adopting Gradient Reverse Layer (GRL) \cite{ganin2015unsupervised, ganin2016domain}, or matching the mean and covariance by Correlation Alignment (CORAL) loss\cite{sun2016deep}.

In domain adaptation for depth estimation, AdaDepth\cite{kundu2018adadepth} utilizes not only pixel-level adversarial training but also feature-level residual reconstruction with $l1$ distance. 
$\mathrm{T}^{2}$Net\cite{zheng2018t2net}
 proposes an end-to-end depth estimation network training both style transfer network and depth estimation network simultaneously. 
 Following this study, DESC\cite{lopez2020desc} additionally uses semantic consistency with pre-trained panoptic segmentation network\cite{kirillov2019panoptic} and GASDA\cite{zhao2019geometry} leverages bidirectional cyclic loss with stereo left-right geometric symmetry suggested in \cite{godard2017unsupervised}.
 
 For semantic segmentation, \cite{hoffman2018cycada, chen2019crdoco} argue that it is effective to align both pixel-level and feature-level distribution for enduring domain shift and \cite{hong2018conditional} proposes a  conditional generator with feature-level residuals and adversarial training.

Unlike previous task-specific domain adaptation algorithms that use unpaired data from different domains, we leverage feature-level mutual information between the synthetic images and their paired realistic ones generated by the style-transfer network as shown in Fig. 1.
It encourages the network to learn the geometric and semantic representation independent of the domain distribution and endure the domain shift.

 \begin{figure*}[t]
    \centering
    \includegraphics[width=0.85\textwidth]{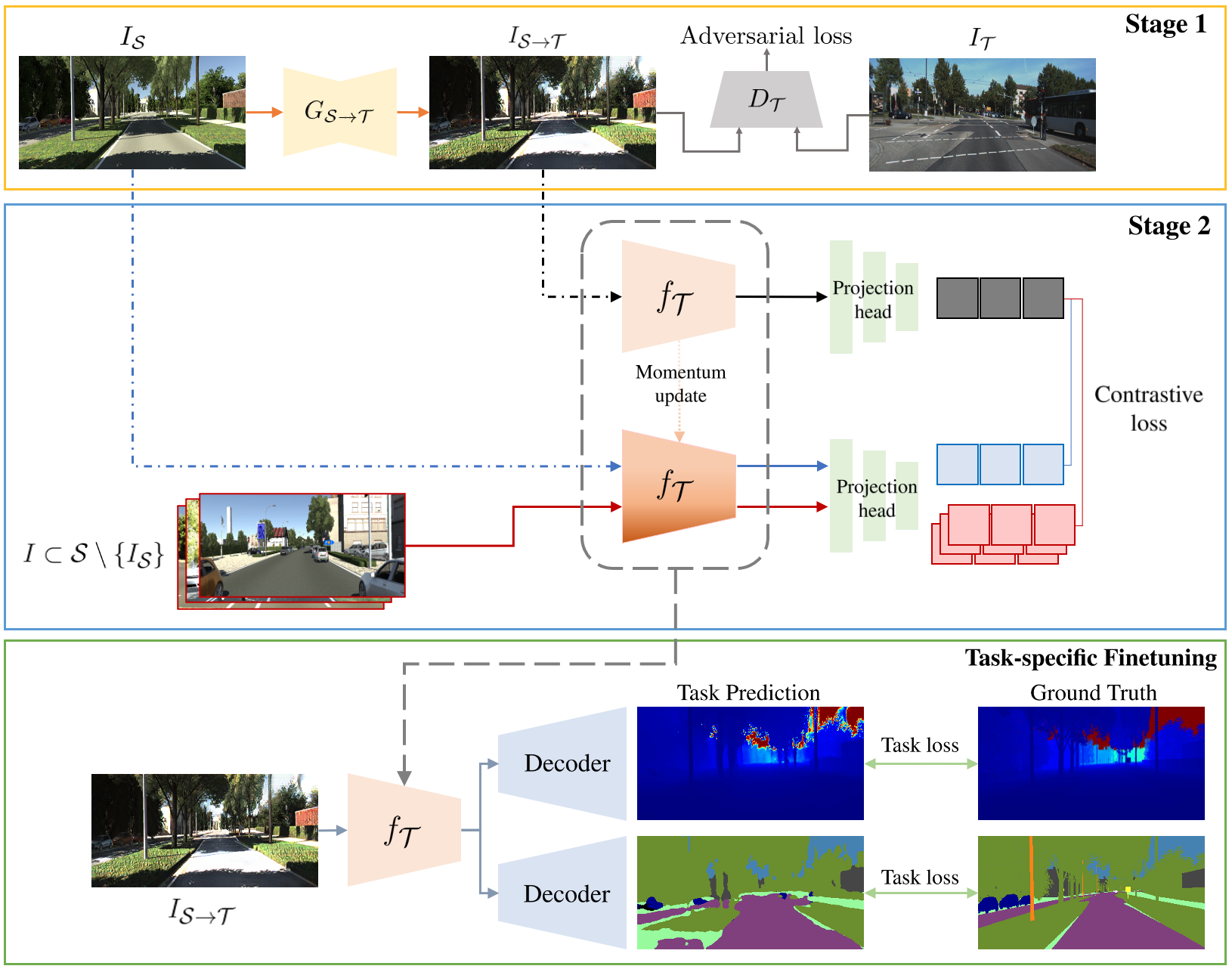}
    \caption{The proposed framework DACL consists of two stages. 
    i) A style transfer network $G_{\mathcal{S}\rightarrow\mathcal{T}}$ is trained with images from source domain $\mathcal{S}$ and target domain $\mathcal{T}$ by adversarial loss. 
    ii) The feature extractor $f_{\mathcal{T}}$ is trained with paired images generated in stage 1 and contrastive loss to learn the domain-agnostic representation. 
    The feature extractor trained in stage 2 $f_{\mathcal{T}}$ is used as the encoder of task-specific network and the entire network is trained with labels from the source domain. 
    For simplicity, we omit the cyclic loss in stage 1 and illustrate only unidirectional flow from the source domain $\mathcal{S}$ to the target domain $\mathcal{T}$ where the entire DACL is a bidirectional algorithm. 
    See Section \RomanNumeralCaps{3} for more details.} 
    \label{fig:algo}
\end{figure*}

\section{Method}
We propose a two-stage unsupervised domain adaptation framework so that the deep neural network can learn the domain-agnostic representation to overcome the performance degradation due to the domain gap between the source domain $\mathcal{S}$ and the target domain $\mathcal{T}$. 
Our proposed Domain-Agnostic Contrastive Learning (DACL) utilizes a bidirectional style transfer algorithm and momentum-based contrastive learning to extract domain-agnostic features such as semantic or structural information which boost the performance of task-specific networks. 
As our DACL aims to train the network in an unsupervised manner within the target domain $\mathcal{T}$, the dataset used during training is composed of images paired with labels from the synthetic source domain $\{x_{s}, y_{s}\} \subset \mathcal{S}$ and images from the realistic target domain without labels $\{x_{t}\} \subset \mathcal{T}$.
\subsection{Stage 1: Bidirectional Style Transfer} 
First, we train two style transfer networks $G_{\mathcal{S}\rightarrow\mathcal{T}}$ and $G_{\mathcal{T}\rightarrow\mathcal{S}}$ to bridge the gap between the source domain $\mathcal{S}$ and the target domain $\mathcal{T}$ in a bidirectional flow. 
As suggested in \cite{goodfellow2014generative}, we optimize the network $G_{\mathcal{S}\rightarrow\mathcal{T}}$ in an adversarial min-max strategy to generate indistinguishable realistic images from the synthetic source image $x_{s}$. Then, the discriminator $D_{
\mathcal{T}}$ is trained to differentiate the generated image $G_{\mathcal{S}\rightarrow\mathcal{T}}(x_{s})$ from the realistic image $x_{t}$ which is sampled from the target domain $\mathcal{T}$ and vice versa for $G_{\mathcal{T}\rightarrow\mathcal{S}}$ and $D_{\mathcal{S}}$.
\begin{equation}
    \begin{split}
    \mathcal{L}_{adv} = \,&\mathbb{E}_{x_{t}\sim \mathcal{T}}[D_{\mathcal{T}}(G_{\mathcal{T}\rightarrow\mathcal{S}}(x_{t}))] \,+\\
    &\mathbb{E}_{x_{s}\sim \mathcal{S}}[1- D_{\mathcal{S}}(x_s)] \,+\\
        &\mathbb{E}_{x_{s}\sim \mathcal{S}}[D_{\mathcal{S}}(G_{\mathcal{S}\rightarrow\mathcal{T}}(x_{s}))] \,+\\
    &\mathbb{E}_{x_{t}\sim \mathcal{T}}[1 - D_{\mathcal{T}}(x_t)]
    \end{split}
\end{equation}
\begin{table*}[t]\small
\centering
\begin{tabular}{c|c|c||cccc|ccc}
\hline

\hline
\multirow{2}{*}{Method} & \multirow{2}{*}{Supervised} & \multirow{2}{*}{Cap} & \multicolumn{4}{c|}{Error Metrics} & \multicolumn{3}{c}{Accuracy Metrics} \\ \cline{4-10}
                  &                   &                   &                     Abs Rel   &  Sq Rel   &  RMSE   & RMSE log   &   $\delta<1.25$    &   $\delta<1.25^2$    &  $\delta<1.25^3$    \\
\hline

    AdaDepth\cite{kundu2018adadepth}              &      No             &             $80m$             & 0.214   & 1.932   & 7.157   &0.295   & 0.665    & 0.882    & 0.950 \\
    AdaDepth\cite{kundu2018adadepth}              &      Semi             &             $80m$             & 0.167   & 1.257   &5.578   &0.237   & 0.771    & 0.922    & 0.971 \\
    $\mathrm{T}^{2}\mathrm{Net}$\cite{zheng2018t2net}      &      No             &             $80m$             & {0.173}   & {1.396}   & {6.041}   &{0.251}   &   {0.757}    & {0.916}    &{0.966} \\
        DESC\cite{lopez2020desc}              &      No             &             $80m$             & 0.156   & 1.067   & 5.628   &0.237   & 0.787    & 0.924    & 0.970 \\
     GASDA \cite{zhao2019geometry}     &      No             &             $80m$             & {0.149}   & {1.003}   & {\bf 4.995}   &{0.227}   &   {0.824}    & {0.941}    &{0.973} \\
\hline

     Ours (GASDA+DACL)     &      No             &             $80m$             & {\bf 0.118}   & {\bf 0.996}   & {5.116}   &{\bf 0.214}   &   {\bf 0.852}    & {\bf 0.944}    &{\bf 0.974} \\
\hline

\hline
    AdaDepth\cite{kundu2018adadepth}              &      No             &             $50m$             &  0.203   &  1.734   &  6.251   &  0.284   &   0.687    &   0.899    &  0.958 \\
    AdaDepth\cite{kundu2018adadepth}              &      Semi             &             $50m$             &  0.162   &  1.041   &  4.344   &  0.225   &   0.784    &   0.930    &  0.974 \\
      $\mathrm{T}^{2}\mathrm{Net}$\cite{zheng2018t2net}              &      No             &             $50m$             &  0.168   &  1.199   &   4.674   &  0.243   &   0.772    &   0.912    &   0.966 \\
              DESC\cite{lopez2020desc}              &      No             &             $50m$             & 0.149   & 0.819   & 4.172  &0.221   & 0.805    & 0.934    & 0.975 \\
     GASDA\cite{zhao2019geometry}      &      No             &             $50m$             &  {0.143}   &  {0.756}   &  {\bf 3.846}  &  {0.217}   &    {0.836}    &   {0.946}    &  {\bf 0.976} \\
  
  \hline
       Ours (GASDA+DACL)      &      No             &             $50m$             &  {\bf 0.112}   &  {\bf 0.724}   &  {3.863}  &  {\bf 0.202}   &    {\bf 0.864}    &   {\bf 0.950}    &  {\bf 0.976} \\
\hline

\hline
\end{tabular}
\caption{Results on KITTI 2015\cite{Menze2015CVPR} using Eigen \textit{et al.}\cite{eigen2014depth} test split for monocular depth estimation. All the methods use vKITTI dataset\cite{Gaidon:Virtual:CVPR2016} as the source domain and KITTI dataset\cite{geiger2012we} as the target domain. Labels from the target domain are disabled during training.}
\label{tb:eigen}
\end{table*}

\begin{figure*}[t]
    \centering
    \includegraphics[width=0.98\textwidth]{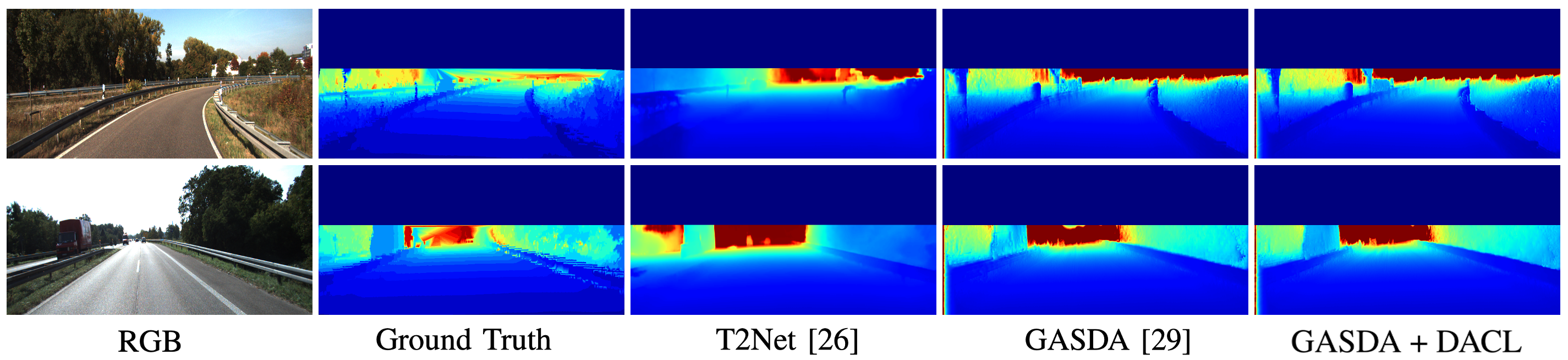}
    \caption{Qualitative results on KITTI dataset\cite{geiger2012we}. The ground truth depths have been interpolated for visualization and all the top regions where the ground truth depths do not exist are masked out.}
    \label{fig:kitti_qual}
\end{figure*}

However, it is difficult to keep meaningful information such as scene geometry and structure of the source image during style transfer due to the inherent instability of adversarial training. 
To remedy such problem, we exploit cycle-consistency loss
\cite{zhu2017unpaired, kim2017learning} so that the style transfer network can map the input image to the other domain while preserving all the important features except for the domain distribution.
\begin{equation}
\begin{split}
    \mathcal{L}_{cyc} =& \,\mathbb{E}_{x_{s}\sim\mathcal{S}}||G_{\mathcal{T}\rightarrow\mathcal{S}}(G_{\mathcal{S}\rightarrow\mathcal{T}}(x_{s})) - x_{s}||_{1} +\\
    &\,\mathbb{E}_{x_{t}\sim\mathcal{T}}||G_{\mathcal{S}\rightarrow\mathcal{T}}(G_{\mathcal{T}\rightarrow\mathcal{S}}(x_{t})) - x_{t}||_{1}
\end{split}
\end{equation}

Since the cycle-consistency loss is insufficient to solve the mode collapse problem during adversarial training, we also utilize an identity reconstruction loss $\mathcal{L}_{idt}$ \cite{taigman2016unsupervised} to strongly regularize the geometric contents during style transfer.
\begin{equation}
\begin{split}
    \mathcal{L}_{idt} =& \,\mathbb{E}_{x_{t}\sim\mathcal{T}}||G_{\mathcal{S}\rightarrow\mathcal{T}}(x_{t}) - x_{t}||_{1} +\\
    &\,\mathbb{E}_{x_{s}\sim\mathcal{S}}||G_{\mathcal{T}\rightarrow\mathcal{S}}(x_{s}) - x_{s}||_{1}
\end{split}
\end{equation}

The full loss function for the style transfer stage $\mathcal{L}_{style}$ is then expressed as:
\begin{equation}
    \mathcal{L}_{style} = \mathcal{L}_{adv} + \lambda_{cyc}\mathcal{L}_{cyc} + \lambda_{idt}\mathcal{L}_{idt},
\end{equation}
where $\lambda_{cyc}$ and $\lambda_{idt}$ are parameters to control the effect of cycle-consistency and identity mapping during training.

\begin{table*}[t]\small
\centering
\begin{tabular}{c|c|c||cccc|ccc}
\hline

\hline
\multirow{2}{*}{Method} & \multirow{2}{*}{Supervised} & \multirow{2}{*}{Dataset} & \multicolumn{4}{c|}{Error Metrics} & \multicolumn{3}{c}{Accuracy Metrics} \\ \cline{4-10}
                  &                   &                &  Abs Rel   &  Sq Rel   &  RMSE   & RMSE log   &   $\delta<1.25$    &   $\delta<1.25^2$    &  $\delta<1.25^3$    \\

\hline
     Monodepth \cite{godard2017unsupervised}&     No              &     K               &  0.124   &  1.388   &  6.125   &  0.217  &  0.841     &  0.936     & 0.975     \\
     Monodepth \cite{godard2017unsupervised}            &     No              &     K+CS             &  0.104   &  1.070   &  5.417   &  0.188  &  0.875     &  0.956     & 0.983     \\
     Atapour {\it et al.}   \cite{atapour2018real}         &      No             &       K + G    &  {0.101}   &  1.048   &  5.308   & 0.184   &   {0.903}    &   {\bf 0.988}    &  {\bf 0.992} \\
     GASDA \cite{zhao2019geometry}&      No             &       K+vK            &  0.106   &  {0.987}   &  {5.215}   & {0.176}   &   0.885    &   0.963    & 0.986 \\
   \hline \hline  
          Ours (GASDA+DACL) &      No             &       K+vK            &  {\bf 0.090}   &  {\bf 0.976}   &  {\bf 4.823}   & {\bf 0.163}   &   {\bf 0.913}    &   0.968    & 0.985 \\

\hline

\hline
\end{tabular}
\caption{Results on $200$ training images of KITTI stereo 2015 benchmark~\cite{geiger2012we, Menze2015CVPR} for monocular depth estimation. For the dataset, K represents real KITTI dataset\cite{geiger2012we}, vK is Virtual KITTI dataset\cite{Gaidon:Virtual:CVPR2016}, CS is CityScapes dataset\cite{Cordts2016Cityscapes} and G is the dataset collected in the GTA5 gaming environment.}
\label{tb:stereo}
\end{table*}

\subsection{Stage 2: Contrastive Representation Learning}

In this stage, we train a feature extractor $f(\cdot)$ that captures the consistent and meaningful visual representation from the image regardless of its domain distribution by contrastive loss.
The objective of contrastive learning is to maximize the mutual information (MI) between the query and the positive pair and to minimize the MI between the query and the negative pairs. 
Therefore, the formulation of proper positive and negative pairs is the most important factor for determining the expressiveness of the representation.

Since a task-specific domain adaptation network leverages not only an image from the real environment $x_{r}$ but also style-transferred synthetic-like image $G_{\mathcal{T}\rightarrow\mathcal{S}}(x_{r})$ during evaluation, our DACL trains two feature extractors, $f_{\mathcal{S}}(\cdot)$ and $f_{\mathcal{T}}(\cdot)$, for the input image from the synthetic domain $\mathcal{S}$ and from the realistic domain $\mathcal{T}$ respectively. The only difference between training those feature extractors with contrastive loss is how to formulate query, positive and negative pairs, so we demonstrate the algorithm only on the target domain feature extractor $f_{\mathcal{T}}(\cdot)$ and later explain how to select contrastive pairs for training the source domain feature extractor $f_{\mathcal{S}}(\cdot)$.

We utilize the style transfer network $G_{\mathcal{S}\rightarrow\mathcal{T}}$  trained in stage 1 to generate data for contrastive learning. A fake target $G_{\mathcal{S}\rightarrow\mathcal{T}}(x_{s})$ is used as a query image and its paired source image $x_{s}$ is used as a positive key image. 
As negative data should be sampled from the same domain of the positive pair so that the network captures domain-agnostic features of the image rather than distinguishing the domain distribution itself, we formulate the negative key pairs with multiple images from the source domain $\mathcal{S}$ except for the source image $x_{s}$.

We adopt momentum-based contrastive learning\cite{he2020momentum} with the non-linear projection head $\psi(\cdot)$\cite{chen2020simple} to be less dependent on the batch size and prevent the feature information loss. Representations from the feature extractor are mapped to the low dimensional latent space $\mathcal{Z}$ with $\psi(\cdot)$ and these latent vectors in $\mathcal{Z}$ are directly used as query and key data to measure contrastive loss:
\begin{equation}
\begin{gathered}
        q = \psi(f_{\mathcal{T}}(G_{\mathcal{S}\rightarrow\mathcal{T}}(x_{s}))),\\
        k_{+} = \psi(f_{\mathcal{T}}(x_{s})),\\
        \{k_{-}\} = \{k|k=\psi(f_{\mathcal{T}}(x)), x\subset \mathcal{S} \setminus x_{s}\}.
\end{gathered}
\label{eq:con}
\end{equation}

 The projection head $\psi(\cdot)$ is composed of a 2-layer fully-connected network followed by a ReLU non-linear activation function. After contrastive learning, $\psi(\cdot)$ is thrown away and only the feature extractor $f_{\mathcal{T}}(\cdot)$ is used for downstream tasks.

We leverage the similarity function between the query and key data as a dot product to consider both magnitude and direction of representation in the latent space $\mathcal{Z}$.
By \cite{oord2018representation}, contrastive loss $
\mathcal{L}_{con}$ can be formulated as a simple (N+1)-way classification objective where N is the number of negative keys $\{k_{-}\}$ and 1 is for a single positive key $k_{+}$. 
With an additional temperature parameter $\tau$ for controlling the concentration of the distribution\cite{wu2018unsupervised}, the objective loss function of contrastive learning can be expressed as below.
\\
\begin{equation}
    \begin{split}
                \mathcal{L}_{con} &= -  \mathrm{log} \, p(q\,| k_{+}, \{k_{-}\}, \tau)\\
        &= -\mathrm{log}\frac{\mathrm{exp}(q \cdot k_{+}/\tau)}{\Sigma_{i = 1}^{\mathrm{N}+ 1}\mathrm{exp}(q \cdot k_{i}/\tau)}
    \end{split}
\end{equation}
\\
The key data $\{k_{+}, k_{-}\}$ are not just discarded after calculating the contrastive loss, but rather stacked in the dynamic dictionary to gather a large number of negative keys. The size of the dictionary is kept the same during training, so the oldest values are removed when the newest key data are enqueued.
Also, each query and key data is used to train two identical feature extractors.
A parameter of feature extractor for the query input $\theta_{q}$ is updated by gradient descent with the loss signals from contrastive loss, but the parameter of feature extractor for the key input $\theta_{k}$ is updated only by momentum for the learning stability:
\begin{equation}
    \theta_{k} \leftarrow m\theta_{k} + (1-m)\theta_{q},
\end{equation}
where m denotes a momentum coefficient.

Similar to training the feature extractor for the target domain $f_{\mathcal{T}}(\cdot)$, we adopt a fake source $G_{\mathcal{T}\rightarrow\mathcal{S}}(x_{t})$ as the query image, image from the realistic target domain $x_{t}$ as a positive image and multiple images from the target domain except for $x_{t}$ as the negative pairs to train the feature extractor for the source domain  $f_{\mathcal{S}}(\cdot)$. 

The feature extractors $f_{\mathcal{S}}(\cdot), \, f_{\mathcal{T}}(\cdot)$ trained via our proposed style transfer and contrastive learning stage are then used as the encoder of the task-specific network. 
The final objective is to reduce the distance between the output of the task-specific network and its paired labels from the source domain $\mathcal{S}$ in a supervised manner. 
Our DACL is applicable to any task-specific network which utilizes the encoder-decoder structure adopted by most networks that generate an output with a large spatial size. The overall framework of DACL is illustrated in Fig. \ref{fig:algo}.

\begin{table*}[htp]\small
\centering
\begin{tabular}{c||c|c|c|c|c|c|c|c|c|c|c|c|c}
\hline

\hline
Method & Car& Building& Grail& Pole& Road& Sky& Terrain& T Light& T Sign& Truck& Van& Veg&mIoU\\
\hline

Source Only&46.5&18.2&{\bf 1.7}&0.0&73.7&68.9&17.7&0.0&0.0&2.0&{\bf 0.2}&63.1&26.0\\
GRL\cite{ganin2015unsupervised}&47.1&18.2&{\bf 1.7}&0.0&75.4&70.5&17.7&0.0&{\bf 5.5}&1.6&0.0&61.8&28.2\\
CycleGAN\cite{zhu2017unpaired}&49.1&16.5&0.0&0.0&75.9&66.6&19.7&0.0&0.0&1.8&0.0&62.2&26.1\\
\hline \hline
DACL&{\bf 51.8}&{\bf 18.3}&0.0&0.0&{\bf 77.1}&{\bf 71.0}&{\bf 28.0}&0.0&0.0&{\bf 2.3}&0.0&{\bf 63.4}&{\bf 30.9}\\
\hline

\hline
\end{tabular}
\caption{Results on 200 training images from KITTI semantic segmentation benchmark\cite{Alhaija2018IJCV} for Virtual KITTI\cite{Gaidon:Virtual:CVPR2016} $\rightarrow$ KITTI\cite{geiger2012we} domain adaptation. Categories in KITTI are different from the training domain Virtual KITTI, so some of them are merged or separated during evaluation.}
\label{tb:seg}
\end{table*}

\begin{figure*}[t]
    \centering
    \includegraphics[width=0.98 \textwidth]{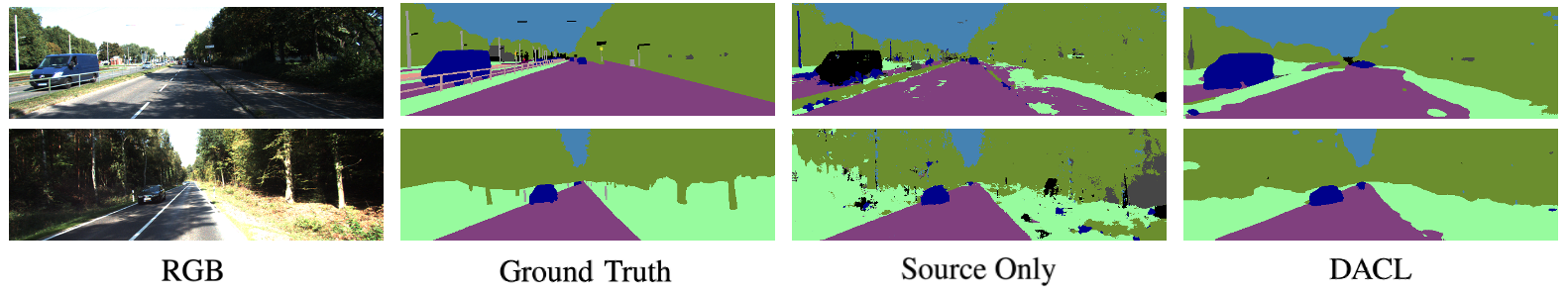}
    \caption{Qualitative results of the semantic segmentation task on training images of the KITTI semantic segmentation benchmark\cite{Alhaija2018IJCV}. The results of DACL are from the unidirectional flow, source to target domain.}
    \label{fig:seg}
\end{figure*}

\section{Experimental Results}

To demonstrate the effectiveness of our proposed domain adaptation framework, we first explain datasets that are used during training and evaluation. 
We then exhibit details of the network architecture and compare our method with existing algorithms in two different tasks which are widely applied for autonomous driving, \textit{i.e.}, monocular depth estimation, and semantic segmentation.

\subsection{Dataset}

Virtual KITTI (vKITTI)\cite{Gaidon:Virtual:CVPR2016} is used as the synthetic source domain dataset which is sampled in the computer-graphical virtual environment that simulates the autonomous driving settings in the real world. 
vKITTI is composed of 21,260 images paired with densely annotated depth maps and semantic segmentation labels. 
We harness KITTI dataset \cite{Menze2015CVPR} as the realistic dataset that is composed of 42,382 rectified images, and some images are paired with human-annotated semantic segmentation labels\cite{Alhaija2018IJCV} and LiDAR point cloud data for guiding the ground truth information. 
In the experiment, we only leverage these labels from KITTI during the evaluation procedure since our proposed domain adaptation method and its comparable previous studies are all trained in unsupervised settings. 
The resolution of both datasets is 375 $\times$ 1242, but the images are resized to 192 $\times$ 640 during training and evaluation for time and computing efficiency.

\subsection{Network Architecture}

Our proposed framework DACL consists of two stages and each stage requires different network architectures according to their objective. 
The architectures for bidirectional generators $G_{\mathcal{S}\rightarrow\mathcal{T}}$, $G_{\mathcal{T}\rightarrow\mathcal{S}}$ and their corresponding discriminators $D_{\mathcal{S}}$, $D_{\mathcal{T}}$ in the style transfer stage are identical to the network provided by CycleGAN\cite{zhu2017unpaired} which utilizes the multiple residual blocks \cite{he2016deep} to increase the connection between hidden layers. Task-specific networks adopt a U-Net\cite{ronneberger2015u}-based structure with skip-connections suggested in \cite{zheng2018t2net}, where the outputs of the networks have 4 different spatial sizes, from the coarse to the fine prediction.
The only difference between the depth estimation network and the semantic segmentation network is the number of the last layers which determines the number of output channels. 
The output of the depth prediction network has a single channel whereas the semantic segmentation prediction has multiple channels, the number of which is equivalent to the categories to classify.

\subsection{Monocular Depth Estimation}

For a fair evaluation, we use the train-test split for KITTI dataset suggested in \cite{eigen2014depth} where 697 test images are sampled from the 29 scenes and 22,600 images from the remaining 32 scenes are adopted for training both the style transfer and depth estimation network. 
We leverage train images and their paired depth labels from vKITTI, and stereo rectified images from KITTI to apply additional geometric consistency loss suggested in GASDA\cite{zhao2019geometry}. 
In TABLE \ref{tb:eigen}, our proposed DACL with stereo matching\cite{godard2017unsupervised, zhao2019geometry} outperforms the previous state-of-the-art domain adaptation approaches for monocular depth estimation in most of the metrics. 
Especially compared to the na\"ive GASDA model, the application of DACL leads to almost 20\% of the performance increase in Abs Rel, and 3\% increase in the accuracy metric. 
In TABLE \ref{tb:stereo}, we evaluate our method with 200 training images of the KITTI stereo 2015 benchmark images paired with corresponding disparity maps.
As \cite{atapour2018real} employed the synthetic data collected in GTA5 where the distribution of the environment is more similar to the real world than vKITTI, it suffers less from performance degradation and achieves better performance in some accuracy metrics. 
Despite leveraging data from the environment with a larger domain gap, our method exceeds other competing methods including \cite{atapour2018real} in all the error metrics and one accuracy metric. 
Some qualitative results of DACL and its comparative methods are shown in Fig. \ref{fig:kitti_qual}. 
Our approach predicts the most similar depth map to the ground truth and preserves structural information of the image. 
DACL generates distinct boundaries between the objects and the background, such as guardrails or vehicles, which indicates more accurate depth map compared to blurry edges shown in comparative methods.

\subsection{Semantic Segmentation}

We also experiment our DACL for semantic segmentation task in a unidirectional flow, from the synthetic domain to the realistic domain, and only utilize two networks, $G_{\mathcal{S}\rightarrow\mathcal{T}}$ and $f_{\mathcal{T}}$, during training and evaluation. 
The network is trained to classify 12 categories with pixel-wise cross entropy loss and no regularization term is added.
In TABLE \ref{tb:seg}, DACL shows better performance compared to previous approaches upon some categories with the mean-Intersection-over-Union (mIoU) metric. Some categories are merged or separated during evaluation as the categories in vKITTI are slightly different from KITTI. 
The evaluation on some categories such as pole or traffic light shows poor performance across all the algorithms and it is due to the bias of the training dataset as the top-5 categories accounts for almost 90\% of the entire area. 
Also from the qualitative results shown in Fig. \ref{fig:seg}, our DACL is much less noisier compared to training only on the source domain dataset and prevents misclassification due to the domain shift.

\section{CONCLUSIONS}

In this paper, we propose a two-stage unsupervised domain adaptation framework called DACL, which trains the ConvNet for multiple autonomous driving tasks with bidirectional style transfer and contrastive representation learning. 
The main motivation of our approach is to learn the domain-agnostic representation and overcome the performance degradation due to the domain shift using the synthetic source images and realistic target images without any supervisory labels from the real world. 
Experiments on KITTI benchmarks show that DACL improves the performance of existing domain adaptation algorithms for depth estimation and semantic segmentation both on qualitative and quantitative metrics. 
Our method can be further applied to a variety of tasks that suffer from the domain generalization issue beyond autonomous driving such as image dehazing or super-resolution.

\bibliographystyle{IEEEtran}
\bibliography{root.bib}

\end{document}